\documentclass[sigconf,nonacm,9pt]{acmart}

\setcopyright{none}

\usepackage{booktabs,tabularx,xcolor}
\newcommand{\findingbox}[2]{\par\smallskip\noindent\fcolorbox{black!35}{black!3}{\parbox{\dimexpr\linewidth-2\fboxsep-2\fboxrule\relax}{\small\textbf{#1}\par\smallskip#2}}\par\smallskip}

\title{Understanding Spike Encoding across IoT Sensing Modalities: An Empirical Study}
\author{Yuxuan Hou}
\affiliation{\institution{Southern University of\\Science and Technology}\city{Shenzhen}\country{China}}
\email{12413104@mail.sustech.edu.cn}

\author{Di Yu}
\affiliation{\institution{Zhejiang University}\city{Hangzhou}\country{China}}
\email{yudi2023@zju.edu.cn}

\author{Linshan Jiang}
\authornote{Corresponding author: \href{mailto:jiangls@sustech.edu.cn}{jiangls@sustech.edu.cn}.}
\affiliation{\institution{Southern University of\\Science and Technology}\city{Shenzhen}\country{China}}
\email{jiangls@sustech.edu.cn}

\author{Jin Zhang}
\affiliation{\institution{Southern University of\\Science and Technology}\city{Shenzhen}\country{China}}
\email{zhangj4@sustech.edu.cn}

\hypersetup{pdfauthor={Yuxuan Hou, Di Yu, Linshan Jiang, Jin Zhang}}

\begin{document}

\begin{abstract}
Spiking neural networks (SNNs) offer a path to energy-efficient sensing on Internet of Things (IoT) devices. The input encoder shapes that opportunity by changing the cues available for recognition and the work required to process them. We study these choices across wireless, inertial, surface electromyographic, and tactile sensing. For wireless activity recognition, interleaving reduces estimated energy for arithmetic in the network layers, latency from the prepared input window to returned scores, and energy on the graphics processing unit and system-on-chip (GPU/SoC) rail, compared with repeating the full window on Jetson AGX Orin. With the same input size at each SNN time step, repeating a compact input can yield higher mean development accuracy than interleaving. The inertial experiments show that higher accuracy can come at higher device cost. Direct input can also be faster and more accurate than an event encoder while using less energy on the measured rail, even when its estimated energy for arithmetic in the network layers is higher. Level crossings improve grasp recognition on a later day for previously observed users. In tactile slip detection, level crossings shift the balance between missed slips and false alarms, and checkpoint selection can reverse which input offers higher slip recall. We use these findings to outline how edge sensing systems can profile encoders against the signal structure, the errors that matter to the task, and the cost of the full inference pipeline.
\end{abstract}


\maketitle
\pagestyle{plain}
\raggedbottom

\section{Introduction}
\label{sec:introduction}

Wearable activity monitors and robotic grippers both require efficient on-device sensing, but rely on different temporal cues. Activity recognition typically uses patterns across a sensing window, whereas slip detection depends on transient changes in sensor signals. Meanwhile, edge devices face resource constraints and heterogeneous execution environments~\cite{stisen2015devices,yao2017deepsense,lane2016deepx}. Spiking neural networks (SNNs) offer a potentially energy-efficient alternative by exploiting sparse neural activity~\cite{davies2018loihi}. However, their effectiveness depends on how sensor signals are encoded, which influences both the temporal information preserved and the resulting computational cost.

Sensor signals can be converted into spike events before inference or directly fed into SNNs with internal spiking dynamics~\cite{mostafa2018temporal,rathi2023diet}. Recent work further explores how input information is distributed across SNN time steps~\cite{sun2026encoding}. These steps determine how a completed sensing window is presented to the network, and several steps can process the same samples. Repeatedly processing the full window increases computational overhead, while selecting or quantizing input components may discard useful sensing information. Encoding itself also incurs processing costs. The resulting trade-off involves both the cues retained and the work needed to expose them to the model.

Recent work has characterized SNN performance and deployment costs across sensing modalities~\cite{du2026benchmarking}. These comparisons provide a useful reference for deciding where spiking models may be effective. Choosing an input encoding for a particular sensing task also requires understanding which structures carry useful information, how the downstream model consumes them, and which errors matter to the application. Two configurations can achieve similar overall accuracy while requiring different amounts of computation or producing different types of errors. A useful comparison therefore connects the input representation to the task requirement and the cost of executing it on the target device.

On ARIL, distributing a window across SNN time steps reduces repeated input and device latency, while separate controls with the same input size at each step favor repeating a compact input over presenting four phases in mean development accuracy. Inertial studies identify accuracy--cost trade-offs and show that arithmetic estimates and measured rail energy can rank configurations differently.

Level crossings improve recognition on a later day for previously observed Ninapro DB6 users, with different responses across participants. On uSkin, evaluations that each hold out one object show changes in missed slips and false alarms, with recall rankings reversing under the two checkpoint rules. Together, these findings motivate comparing input coverage, precision, and presentation across SNN time steps under the task's recognition and device-cost requirements. The best measured configurations that meet these requirements provide static baselines for testing runtime refinement or early stopping, including their decision and execution costs.

\noindent Our contributions are as follows.
\begin{itemize}
\setlength{\itemsep}{2pt}
\item We examine how input encoding changes the information a model receives, recognition errors, and device cost across four sensing modalities.
\item We compare continuous, sliced, hybrid, and event inputs across six tasks in wireless, inertial, surface electromyographic (sEMG), and tactile sensing, combining task metrics, arithmetic estimates, and Jetson AGX Orin measurements.
\item We find that reducing repeated input can lower device cost, overall accuracy can conceal differences across users or missed slips, and arithmetic estimates and device measurements can rank encodings differently. These findings guide encoding selection.
\end{itemize}

\section{Encoding and Evaluation}
\label{sec:basis}

We compare configurations defined by their input representation, model, checkpoint, and inference implementation. Each operating point pairs a configuration with its recognition results and measured cost. Input coverage, precision, and SNN time steps are varied under specified users, sessions, objects, and task-error requirements. Figure~\ref{fig:comparison} shows the input choices and the scope of each cost measurement.

\subsection{Input representations and tasks}
\label{sec:arms}

\begin{figure}[t]
\centering
\includegraphics[width=0.95\columnwidth]{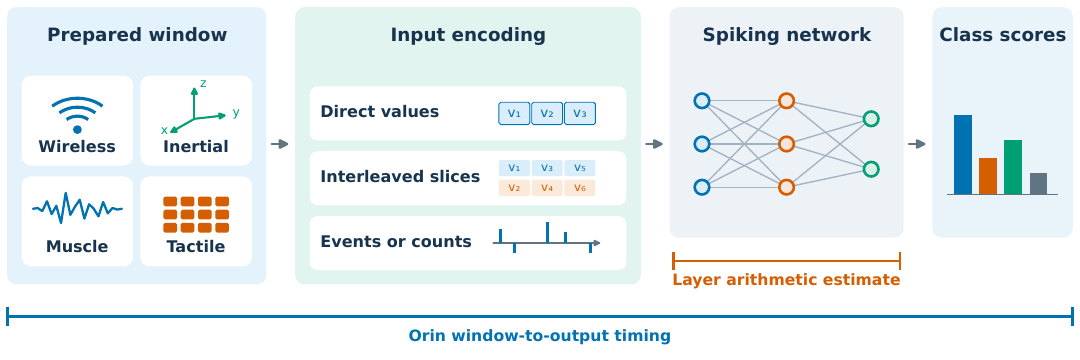}
\caption{Input encoding choices and measurement boundaries for edge sensing. Each workload starts with a prepared window from one modality. The arithmetic estimate covers convolutional and linear operations within the network. Orin replay starts with the prepared host window and ends when classification scores return to the host.}
\Description{A four-stage flow links a prepared sensor window, input encoding, a spiking network, and class scores. Input choices include direct values, interleaved slices, and events or counts. The arithmetic estimate spans only the network, while the timing measurement spans the full flow.}
\label{fig:comparison}
\end{figure}

Each comparison starts with a completed sensing window. SNN time steps can repeat its samples, partition them, or present changes, varying input and inference costs at fixed window duration (Figure~\ref{fig:comparison}).

\textbf{Direct} supplies continuous values to a network with internal spiking dynamics. \textbf{Interleaved slicing} distributes sample positions across steps; \textbf{contiguous slicing} assigns consecutive segments. For UCI-HAR, \textbf{fixed Haar events} use preset subband quotas, whereas the \textbf{gated adapter} draws on Gated Attention Coding~\cite{qiu2024gac} to adapt the input. For MHEALTH, \textbf{cached weighted Haar} uses subband weights learned during training and held constant during inference. The wireless hybrid combines continuous coarse structure with detail events. \textbf{Level crossings} send an initial quantized level and signed changes, including count magnitudes above one when several levels are crossed between samples. The muscle-signal plot labels this encoding Delta. A \textbf{continuous ANN} provides a network reference where available; ANN describes the network rather than an input encoding.

\subsection{Sensing Modalities and Tasks}
\label{sec:protocol}

Each comparison uses public recordings and the same evaluation examples for its input variants. Training seeds vary fitted models; replay blocks measure frozen models. We report seed means, user-specific accuracy on a later day, and tactile results by trial and slip state. Tactile windows overlap, and both checkpoint rules use the same evaluation windows.

\paragraph{ARIL~\cite{wang2019aril}} We split the official training set into 1,004 optimization and 112 development windows. Each comparison uses five seeds and four SNN steps. Full-window and slicing models are trained separately from the equal-size repetition controls; device replay uses one frozen model per input from the slicing comparison.

\paragraph{Widar3.0~\cite{zhang2022widar}} On 22-class body-coordinate velocity profiles, three paired seeds compare Direct with continuous coarse structure plus detail events at epoch 100.

\paragraph{MHEALTH~\cite{banos2014mhealth}} Motion channels from six subjects train the model, and two subjects provide development data. Three seeds compare Direct with cached Haar weighting, including Orin latency and GPU/SoC rail energy.

\paragraph{UCI-HAR~\cite{reyes2013uci}} Participants within the official training set are separated for optimization and development. The laboratory comparison uses three seeds. Orin replays four models from one seed at epoch 100, with eight blocks per model and 64 fixed timing windows.

\paragraph{Ninapro DB6~\cite{palermo2017repeatability}} One paired seed trains on days 1--2 and evaluates day 3 for the same ten users at epoch 40. We compare Direct, an amplitude summary, and level crossings; matched device costs are reported separately.

\paragraph{uSkin~\cite{zenha2021slip}} Each protocol holds out 30 trials of one object. Screwdriver uses three seeds; brush and SpoolSolder each use one paired seed. We compare Direct and level crossings at epoch 40 and at checkpoints chosen on selection trials; device costs use epoch-40 models.

\subsection{Arithmetic energy and device cost}
\label{sec:cost}

NeuroRadar estimates energy from neuron updates and synaptic activity under a Loihi model~\cite{zheng2023neuroradar}. We separately report estimates for arithmetic in the network layers and device measurements~\cite{yik2025neurobench}. In matched replay, active binary inputs contribute one accumulation (AC) per effective connection; continuous inputs contribute multiply--accumulates (MACs) from layer geometry. Summing over steps and averaging over windows gives
\[
E_{\mathrm{arith}}=(0.9\,\mathrm{pJ})N_{\mathrm{AC}}+(4.6\,\mathrm{pJ})N_{\mathrm{MAC}}.
\]
The coefficients describe 32-bit floating-point arithmetic at 45\,nm and 0.9\,V; a MAC combines a 3.7\,pJ multiplication and 0.9\,pJ addition~\cite{horowitz2014energy}. Memory, transfers, background power, encoding scalar operations, and updates to neuron states lie outside the estimate~\cite{sze2017energy}. 

Jetson AGX Orin replay starts with a prepared host window and includes input preparation, encoding, transfers, state reset, inference, and returned scores. We integrate timestamped power on each named supply rail and divide by completed windows, requiring at least 99\% block coverage and no gap above 0.5\,s. Rail measurements include background draw; whole-board input energy is unavailable. UCI-HAR additionally reports CPU/CV and system-5V rails. Appendix~\ref{app:matched-cost-supplement} specifies the models and window roles for the additional matched replays.

\section{Findings by Sensing Modality}
\label{sec:lessons}

\subsection{Wireless Activity and Gesture Recognition}
\label{sec:wireless}

\findingbox{Finding 1. On ARIL, interleaving reduces device cost relative to full-window repetition. At the same input size per step, compact repetition can yield higher mean development accuracy than interleaving.}{}

\begin{figure}[t]
\centering
\includegraphics[width=0.82\columnwidth]{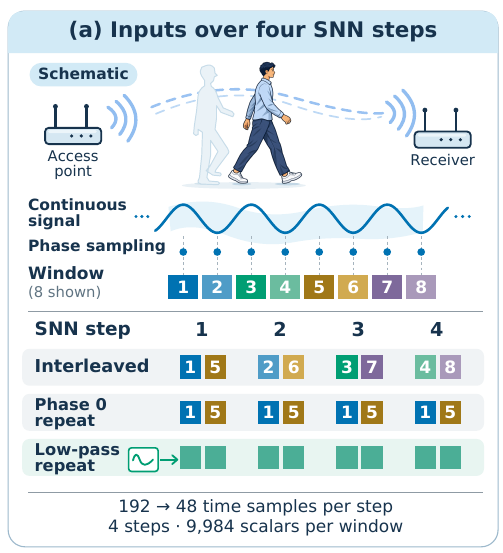}
\caption{ARIL input layouts over four SNN steps. Accuracy and arithmetic results are shown separately in Figure~\ref{fig:wireless-results}.}
\Description{Original scene and phase layout for interleaving, fixed-phase repetition and low-pass phase repetition.}
\label{fig:wireless}
\end{figure}

\begin{figure}[t]
\centering
\includegraphics[width=0.95\columnwidth]{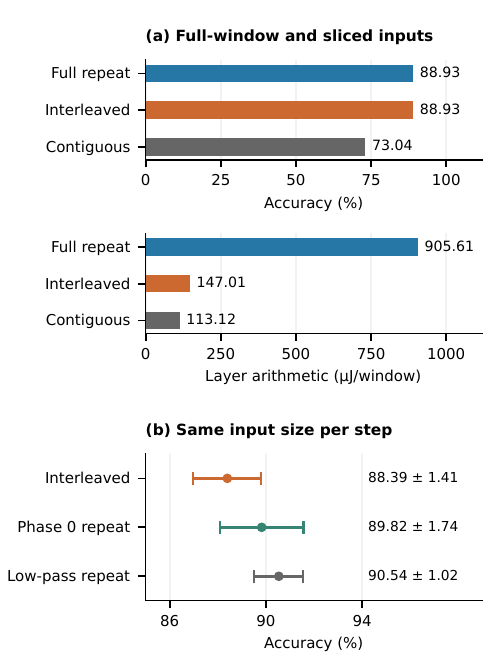}
\caption{ARIL results. (a) Full windows versus slices: five-seed accuracy and layer arithmetic. (b) Equal input size per step: mean and sample standard deviation of the highest development accuracy over five paired seeds trained for 60 epochs. The two panels use separately trained cohorts; device measurements use the full-window/slicing cohort.}
\Description{Conventional accuracy and arithmetic plots, plus matched-control means and error bars.}
\label{fig:wireless-results}
\end{figure}

\paragraph{Input presentation across time steps}
ARIL Wi-Fi channel measurements are linearly interpolated to 192 time points~\cite{wang2019aril}. We split the official training set by class into 1,004 optimization and 112 development windows. Full windows and slices use the SpikingVGG5 backbone family and four SNN time steps. Direct repeats full windows after min--max scaling; unscaled contiguous and interleaved slices group adjacent segments or time indices modulo four. Slicing presents all 9,984 values once; Direct presents 39,936.

\paragraph{Recognition and device cost of slicing}
Over five seeds trained for 60 epochs, Direct and interleaving both reach 88.93\% mean accuracy at checkpoints selected on the development set; contiguous slicing reaches 73.04\%. Interleaving reduces estimated energy for layer arithmetic by 83.77\% (Figure~\ref{fig:wireless-results}a). In an Orin replay of one frozen model per input from this comparison, Direct and interleaving have median latencies of 26.50 and 13.09\,ms and GPU/SoC rail energies of 725.98 and 121.64\,mJ per window. Both slicing layouts increase CPU/vision rail energy. This comparison changes input geometry, scaling, and internal activity; its device benefit concerns the measured inference path and rail. Appendix~\ref{app:extra-evidence} provides layer counts and the protocol with eight replay blocks.

\paragraph{A compact repeated input can be sufficient}
To vary the amount of distinct input while keeping the input size at each step fixed, we train a separate cohort of models using interleaving, repetition of a fixed phase, or repetition after low-pass filtering for 60 epochs over five paired seeds on the same Orin backend. They share unscaled input, optimization and development windows, initializations, sample orders, and four inputs of shape $1\times52\times48$. Interleaving presents four phases; repetition of a fixed phase presents positions $0,4,\ldots,188$ at every step; the low-pass control filters the completed window and repeats the retained positions. The fixed 33-tap filter allows all original positions to influence its retained outputs. Mean accuracies at checkpoints selected on the development set are $88.39\pm1.41\%$, $89.82\pm1.74\%$, and $90.54\pm1.02\%$, respectively (mean and sample standard deviation). Table~\ref{tab:aril-new-controls} reports the comparison at the final epoch; Appendix~\ref{app:extra-evidence} gives endpoint selection and filter work.

\begin{figure*}[t]
\centering
\includegraphics[width=0.92\textwidth]{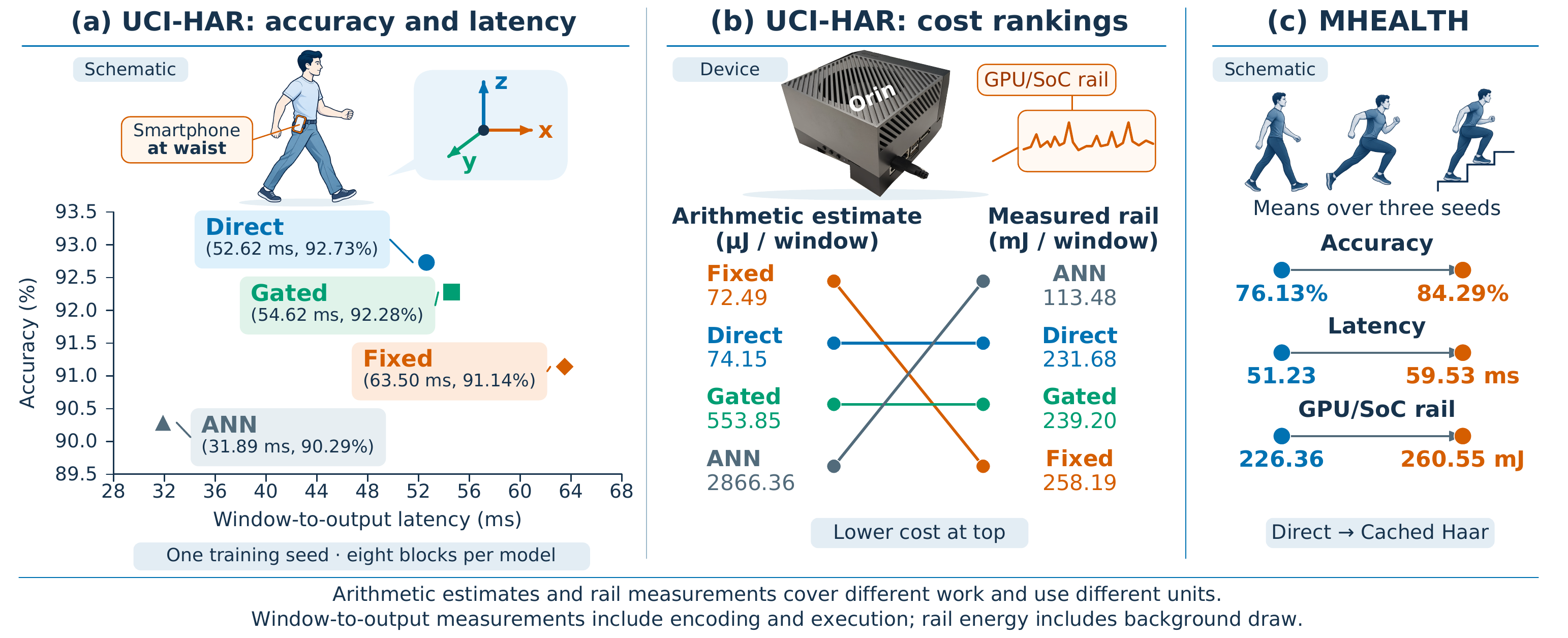}
\caption{Inertial accuracy and inference cost. MHEALTH shows means of per-model eight-block medians over three training seeds at epoch 20. UCI-HAR replays four frozen models trained with one seed over 32 blocks (eight per model). Lower estimates for layer arithmetic can accompany higher GPU/SoC rail energy on the dense Orin backend. Rail measurements follow Section~\ref{sec:cost}.}
\Description{A person wears a smartphone at the waist, and an Orin board illustrates power measurement on an internal supply rail. On MHEALTH, higher accuracy comes with higher measured latency and rail energy. On UCI-HAR, fixed event encoding has a slightly lower arithmetic estimate than Direct but higher measured rail energy. The ANN has the highest arithmetic estimate and the lowest rail energy, with lower accuracy.}
\label{fig:inertial}
\end{figure*}

\paragraph{State and phase coverage}
Interleaving exceeds contiguous slicing at the same compact shape. Repetition of a fixed phase has higher selected accuracy in three paired seeds and lower accuracy in two; repetition after low-pass filtering is higher in four and tied in one. For the five interleaved models trained for the comparison with repeated full windows, resetting state between steps reduces mean accuracy to 25.18\%, establishing their state dependence. The repetition controls separately show that high development accuracy can be obtained by repeating one phase; their device costs remain unmeasured. These controls separate the amount of distinct input from the number of times it is processed.

\paragraph{Continuous coarse velocity structure}
On 22-class Widar3.0 body-coordinate velocity profiles (BVPs)~\cite{zhang2022widar}, a hybrid keeps the coarse velocity structure continuous and adds selected detail events. With the same spiking attention backbone running for four steps, it raises development accuracy at epoch 100 from 76.47\% to 77.69\% over three paired seeds. In an ablation using one seed and holding the rule for selecting events fixed, preserving the coarse component reaches 77.27\%, versus 73.46\% when all bands are represented by quantized events. These comparisons motivate retaining useful coarse structure while testing which detail components can be represented as events. On Orin, replay of one pair of epoch-100 models yields 76.58\% and 77.41\% accuracy for Direct and the hybrid, with paired increases of 12.48\,ms and 40.03\,mJ per window on the GPU/SoC rail; this single-seed device comparison complements the three-seed recognition result (Table~\ref{tab:additional-cost}).

\subsection{Inertial Activity Recognition}
\label{sec:inertial}

\findingbox{Finding 2. On the dense Orin backend, a lower estimate for layer arithmetic can accompany higher measured GPU/SoC rail energy.}{}

\paragraph{MHEALTH exchanges device cost for accuracy}
We use MHEALTH motion channels~\cite{banos2014mhealth}, with six subjects for training and two for development. The cached weighted Haar encoder ranks signed residual events using four subband weights learned during training and fixed during inference, retaining up to 25\% of coefficient positions. It computes the constant weights once and reuses them while preserving the encoding and predictions. This implementation raises development accuracy from 76.13\% to 84.29\% over three training seeds relative to Direct. In the matched Orin replay, latency from the prepared window to returned scores rises from 51.23 to 59.53\,ms, and GPU/SoC rail energy from 226.36 to 260.55\,mJ per window (Figure~\ref{fig:inertial}c). These points average each model's eight-block median over three seeds. The mean within-round paired increases are 8.24\,ms and 34.18\,mJ per window, respectively, with the same direction in all three seeds. The effective network accumulation count also increases; this result quantifies the device cost of the recognition gain.

\paragraph{UCI-HAR favors a simpler input}
UCI-HAR uses smartphone inertial recordings~\cite{reyes2013uci}. Using different participants for optimization and development within its official training set, Direct and fixed Haar events reach 93.20\% and 91.05\% mean development accuracy over three training seeds. The Orin comparison replays four models saved at epoch 100 from one prespecified seed; SNNs use four steps, and the ANN uses the frozen recipe with continuous activations.

For Direct SNN, fixed Haar events, the gated adapter, and continuous ANN, median latencies are 52.62, 63.50, 54.62, and 31.89\,ms, and their GPU/SoC rail energies are 231.68, 258.19, 239.20, and 113.48\,mJ per window (Figure~\ref{fig:inertial}). The fixed Haar event encoder is slower and uses more energy on this rail at lower accuracy than Direct. The gated adapter also takes slightly longer in this run. The ANN trades 2.44 percentage points of accuracy for 39.40\% lower median latency and 51.02\% lower GPU/SoC rail energy relative to Direct SNN. The task's accuracy requirement determines which operating points are feasible.

\paragraph{Arithmetic and measured energy can rank inputs differently}
For these same frozen models, estimated energy for arithmetic in the network layers is 74.15, 72.49, 553.85, and 2,866.36\,$\mu$J per window. Fixed Haar events has a slightly lower estimate than Direct, yet higher measured rail energy. The ANN has the highest arithmetic estimate and the lowest measured rail energy. The reference arithmetic model prices logical accumulations for binary inputs cheaply; the dense GPU backend executes them through tensor operations and also incurs frontend, state, memory, and background costs. Layer arithmetic and device measurements describe different parts of the cost; the measured operating points determine feasibility on this backend.

Each model uses eight counterbalanced replay blocks on 64 fixed windows and the timing boundary in Section~\ref{sec:cost}. Appendix~\ref{app:extra-evidence} and Table~\ref{tab:uci-matched-cost} provide tail latencies and individual rail measurements. These measured profiles allow an application to select an operating point that meets its recognition and cost requirements.

\subsection{Surface Electromyographic Grasp Recognition}
\label{sec:semg}

\findingbox{Finding 3. On Ninapro DB6, level crossings can improve grasp recognition for previously observed users evaluated on day 3.}{}

\begin{figure}[t]
\centering
\includegraphics[width=0.95\columnwidth,trim=0 9bp 0 0,clip]{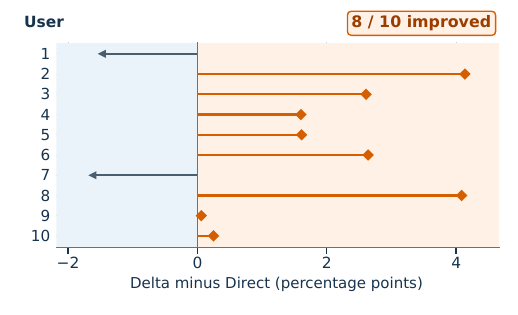}
\caption{Change in accuracy on day 3 from level crossings (Delta) relative to Direct for ten previously observed Ninapro DB6 users. One paired seed is evaluated at epoch 40. The mean gain across users is 1.39 percentage points, with eight users improving.}
\Description{A horizontal difference plot shows one accuracy difference per participant, with the Direct baseline marked by a vertical zero line. Users 1 and 7 fall left of zero. The other eight users fall right of zero.}
\label{fig:semg}
\end{figure}

\paragraph{Recognition across recording days}
On Ninapro DB6, we train a pooled model on days 1 and 2 and evaluate day 3 for the same ten users~\cite{palermo2017repeatability}. Each window contains 400 samples at 2\,kHz from 14 electrodes, advances every 50\,ms, and takes its final sample's movement label. The development split contains 285,696 overlapping windows. We average accuracies computed separately for each user.

At epoch 40, we compare three inputs to a one-dimensional residual spiking backbone under matched initialization and training order. Direct passes standardized samples; the amplitude control adds a causal 20\,ms root-mean-square envelope; level crossings send the initial quantized level and signed changes whose cumulative sum recovers the quantized signal.

\paragraph{The gain appears across users, with a range of responses}
Level crossings raise accuracy averaged across users from 46.98\% to 48.36\% and macro-F1 (mean class F1) from 38.19\% to 40.07\%. The gain is 1.39 percentage points (95\% paired bootstrap interval obtained by resampling users: 0.17 to 2.59 for this seed). The amplitude control reaches 44.88\%. Eight users improve; two lose about one and a half points (Figure~\ref{fig:semg}). This breakdown shows whether a mean gain is shared across intended users. Signed counts enter a dense first convolution. An energy estimate should include this convolution as well as the later spiking layers. Orin replay of the same epoch-40 weights on 80 prespecified day-3 windows adds 0.141\,ms and 0.457\,mJ per window on the GPU/SoC rail relative to Direct, using within-round paired means (Table~\ref{tab:additional-cost}).

\subsection{Tactile Slip Detection}
\label{sec:tactile}

\findingbox{Finding 4. Tactile encoding trades missed slips against false alarms, and checkpoint selection can reverse the slip-recall ranking.}{}

\paragraph{Object-held-out slip detection}
Slip detection can provide feedback for stabilizing a grasp~\cite{james2021slip}. We form 36-sample windows from uSkin Hall-effect readings and use the last sample's label~\cite{zenha2021slip}. For each object held out, we train separate Direct models and models that use level crossings, fitting preprocessing on the optimization trials. The screwdriver protocol uses 48 optimization and 12 selection trials, with three seeds evaluated on 30 held-out trials and 9,703 windows. The brush and SpoolSolder protocols each use one paired seed and 30 held-out trials, yielding 4,106 and 9,890 windows. Checkpoints are fixed before evaluating the held-out trials. The continuous ANN reference belongs to the screwdriver protocol.

\begin{figure}[t]
\centering
\includegraphics[width=0.92\columnwidth]{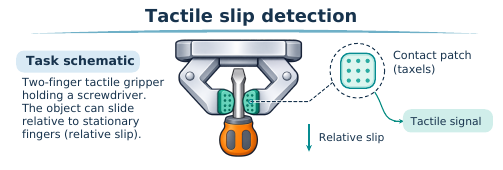}
\caption{Tactile slip sensing. Opposed tactile pads hold a screwdriver; the contact patch senses relative motion along the object.}
\Description{A gripper, contact-patch detail and relative-slip arrow illustrate the sensing task.}
\label{fig:tactile}
\end{figure}

\begin{figure}[t]
\centering
\includegraphics[width=0.95\columnwidth]{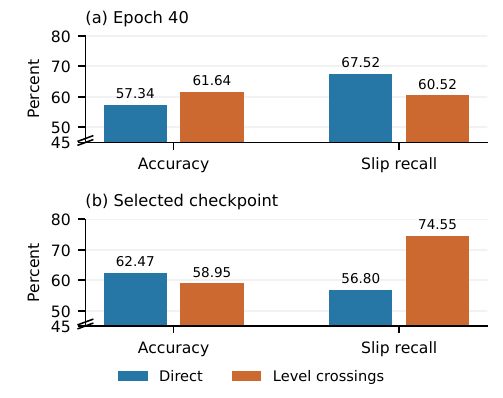}
\caption{Tactile accuracy and slip recall under two checkpoint rules. Means are over three training seeds on the same 30 held-out screwdriver trials. At epoch 40, level crossings raise accuracy and lower recall; selecting the checkpoint with the lowest cross-entropy on selection trials reverses both rankings. Both panels use a truncated vertical scale of 45--80\%.}
\Description{Standard result axes compare Direct and level crossings at epoch 40 and the selected checkpoints.}
\label{fig:tactile-results}
\end{figure}

\paragraph{Accuracy can conceal different rates of missed slips}
At epoch 40, level crossings raise mean accuracy in the screwdriver protocol from 57.34\% to 61.64\%, while slip recall falls from 67.52\% to 60.52\% and false-positive rate falls from 48.14\% to 37.75\% (Figure~\ref{fig:tactile-results}a). The continuous ANN reaches 66.74\% accuracy and 55.37\% recall. On brush, accuracy rises by 11.64 percentage points while recall falls by 2.45 points. On SpoolSolder, accuracy is 1.16 points lower and slip recall is 23.00 points lower for level crossings than for Direct. These outcomes in each protocol show why overall accuracy should be read alongside missed slips and false alarms.

\paragraph{Checkpoint selection changes the error trade-off}
At checkpoints chosen by the lowest cross-entropy on the selection trials, slip recall for level crossings exceeds that of Direct by $+17.75$, $+6.23$, and $+24.63$ points for screwdriver, brush, and SpoolSolder, respectively, accompanied by higher false-positive rates. Encoding and checkpoint selection therefore define the operating point together. At epoch 40, slip recall in the stationary condition is also lower in all three protocols, by 23.94, 8.29, and 28.32 points. Table~\ref{tab:tactile-new-objects} gives brush and SpoolSolder results at both checkpoints; Appendix~\ref{app:tactile-object-protocols} gives errors by slip state.

\paragraph{Logical sparsity and device cost}
At epoch 40, level crossings reduce effective binary accumulations by 37--40\% in the three-seed screwdriver models, yet their paired mean latency and GPU/SoC rail energy increase by 0.235\,ms and 0.813\,mJ per window. These selection-window replays retain the same dense layer geometry and neuron-state sizes. Brush and SpoolSolder show the same separation between fewer logical accumulations and measured execution cost (Table~\ref{tab:additional-cost}).

\paragraph{Encoder selection under slip-detection requirements}
For a gripper, set the required slip recall and acceptable false-positive rate, then select the input and checkpoint on the selection trials. Evaluate the resulting operating points on held-out object trials. These measurements describe detection over input windows; a closed-loop test can connect physical slip onset to corrective action.

\section{Discussion and Open Questions}
\label{sec:vision}

The experiments support selecting an encoder and checkpoint under the intended recognition and device requirements. Three questions remain for extending these measured configurations to runtime adaptation.

\paragraph{Which information is sufficient for the intended use?}
Encoder comparisons should specify the users, sessions, objects, and task errors used to judge recognition. ARIL fixes the number of SNN time steps while varying the distinct input presented. A further comparison could vary input content and repeated processing independently to determine when additional samples or state updates improve recognition.

\paragraph{Which operating points are feasible on the device?}
The MHEALTH and UCI-HAR comparisons include preparation, encoding, transfers, state reset, inference, and returned scores within the same timing scope. An application can retain configurations that meet its recognition and latency or energy requirements. Interoperable model descriptions can then support comparisons across runtimes~\cite{pedersen2024nir}. On interfaces accepting only spikes, the comparison must also include conversion of a continuous input into events.

\paragraph{When does adaptation repay its overhead?}
A runtime policy could request more input, refine precision, continue computing, or stop as input conditions or application requirements change. It should be compared with the best measured static configuration that meets the same recognition requirement for the same users, sessions, or objects. The comparison should include the costs of the policy decision, preparation, transfers, and inference.

\section{Related Work}
\label{sec:related}

\paragraph{Input representations and temporal organization}
Spiking models can receive hardware events or encoded samples. NeuroRadar generates spikes in the analog frontend; other wireless systems encode Doppler maps, process raw radar, or perform transforms and classification on neuromorphic hardware~\cite{zheng2023neuroradar,tsang2021radar,arsalan2022adc,lopez2022fourier,stuijt2021mubrain}. Preprocessing and hybrid Wi-Fi comparisons highlight the roles of representation and architecture~\cite{safa2023radar,lu2026sws,yang2023sensefi}. Inertial work studies spatiotemporal models, continuous input, and sensors that produce spikes~\cite{li2022wearable,fra2022har,fra2024lmu,jiang2023motion}. Du et al.\ compare encodings with Direct as a continuous input reference~\cite{du2026benchmarking}. Our ARIL controls compare complementary phases with a compact input repeated over the same number of SNN time steps.

\paragraph{Use conditions and application errors}
Studies of surface electromyography use event and delta representations for hand-gesture recognition~\cite{donati2019emg,ma2020emg,ceolini2020fusion,ding2026emg} and adapt to changes in how muscle activity is measured~\cite{moin2021wearable,guo2024spgesture}. Tactile research includes touch sensors that produce spikes and asynchronous electronic skin~\cite{lee2019aces,kim2018afferent,chun2021tskin,wardcherrier2020neurotac}, event-based recognition, and slip detection~\cite{braille2022,taunyazov2020vt,lu2026slip,zhao2025spiketouch}. We examine recognition for each user on a later day and distinguish overall accuracy from slip recall to identify encoding differences that matter to the application.

\paragraph{Execution on sensing hardware}
Bian et al.\ directly compare inertial encodings on Loihi~2, complementing wrist-worn delta coding and variable-precision execution~\cite{bian2024encoding,bian2023iswc,leto2026precision}. Other work deploys muscle-signal models on low-power hardware~\cite{bezugam2023loihi,scrugli2025semg}. NeuroBench and interoperable representations support comparisons across neuromorphic workloads and runtimes~\cite{yik2025neurobench,pedersen2024nir}. Our device studies pair recognition with latency for the complete inference pipeline and, where measured, energy on individual supply rails, connecting input choices to the cost of executing them.

\section{Conclusion and Future Directions}
\label{sec:conclusion}

Our experiments across four IoT sensing modalities support selecting encodings and checkpoints according to the information needed for recognition, application requirements, and measured device cost.

These findings motivate systems that adjust input coverage, precision, and SNN steps as conditions and budgets change. The open question is when adaptation improves on the best measured static configuration after control and execution costs are included. Answering it would guide how edge devices allocate limited resources to reliable sensing decisions.

\appendix
\makeatletter
\@ACM@balancefalse
\makeatother
\section{Supporting Comparisons}

\subsection{Input comparisons and device measurements}
\label{app:extra-evidence}

\paragraph{ARIL input comparisons}
Both comparisons split each class of the official training set into 1,004 optimization and 112 development windows and use five training seeds. The models comparing full windows with slices are trained separately from those comparing repetition with interleaving at the same input size per step. Without min--max scaling, repeated Direct reaches 78.57\% mean development accuracy in the former comparison. Table~\ref{tab:aril-new-controls} reports selected checkpoints and epoch 60 for the latter.

\begin{table}[H]
\caption{ARIL repetition controls on the same 112 development windows: five paired seeds, 60 epochs per fit, and the same Orin backend. The selected checkpoint has the highest development accuracy, using the earliest epoch in a tie. Entries are mean and sample standard deviation in percent; epoch 60 is shown separately.}
\label{tab:aril-new-controls}
\centering\small
\setlength{\tabcolsep}{3pt}
\begin{tabular}{@{}lrr@{}}
\toprule
Input & \shortstack{Best dev.\\accuracy} & \shortstack{Epoch 60\\accuracy} \\
\midrule
Four-phase interleaving & $88.393\pm1.412$ & $84.286\pm1.494$ \\
Fixed phase repeated & $89.821\pm1.740$ & $86.250\pm1.740$ \\
Low-pass phase repeated & $90.536\pm1.018$ & $85.357\pm1.956$ \\
\bottomrule
\end{tabular}
\end{table}

The 33-tap low-pass filter acts on the completed window before retaining 48 phase-0 outputs, so all original sample positions can influence them. Its frontend work is separate from the layer arithmetic estimate. These comparisons use interpolated ARIL windows.

\paragraph{ARIL device measurements}
Table~\ref{tab:aril-matched-cost} pairs arithmetic counts with device measurements for the full-window and slicing models trained with seed~41. Each layout uses eight counterbalanced 30-second blocks on the same 112 development windows. Direct, interleaving, and contiguous slicing correctly classify 98, 97, and 79 windows. Latencies are medians of block medians. Both slicing layouts increase energy on the CPU/vision rail (\texttt{VDD\_CPU\_CV}). The separately trained repetition controls are not included in this replay.

\begin{table}[H]
\caption{ARIL workload and device cost for frozen models trained with seed 41, evaluated on the same 112 development windows with four SNN time steps and eight replay blocks per layout. MACs and effective ACs are millions per window. Arithmetic covers convolutional and linear layers; latency spans a prepared host window to returned scores. GPU/SoC rail energy follows Section~\ref{sec:cost}.}
\label{tab:aril-matched-cost}
\small
\centering
\setlength{\tabcolsep}{3pt}
\begin{tabular}{@{}lrrr@{}}
\toprule
Metric & \shortstack{Full\\Direct} & Interleaved & Contiguous \\
\midrule
Presented values & 39,936 & 9,984 & 9,984 \\
MACs (M) & 23.015 & 5.763 & 5.763 \\
ACs (M) & 858.071 & 128.683 & 91.968 \\
Arithmetic ($\mu$J) & 878.14 & 142.32 & 109.28 \\
p50 latency (ms) & 26.495 & 13.093 & 12.841 \\
p95 latency (ms) & 27.538 & 13.368 & 13.133 \\
GPU/SoC rail (mJ) & 725.98 & 121.64 & 119.65 \\
\bottomrule
\end{tabular}
\end{table}

\paragraph{UCI-HAR device measurements}
The four frozen models run on PyTorch 2.8 and CUDA 12.6 with 32-bit floating-point computation. Each uses eight counterbalanced blocks, 20 warm-up windows and at least 30 seconds of measurement per block on 64 fixed windows. We compute latency percentiles within blocks and report their medians. All three supply rails meet the telemetry checks in Section~\ref{sec:cost}. Timing starts after window formation and raw-signal standardization and includes transfers, input preparation, encoding, state reset, inference, and returned scores. Table~\ref{tab:uci-matched-cost} reports accuracy, tail latency, and energy for each rail.

\begin{table}[H]
\caption{UCI-HAR frozen models trained with seed 27111 and saved at epoch 100. Accuracy uses all 1,761 development windows; timing uses 64 fixed windows. Each cost entry is the median of eight block summaries. Energy is reported for the individual supply rails defined in Section~\ref{sec:cost}.}
\label{tab:uci-matched-cost}
\centering\small
\setlength{\tabcolsep}{3pt}
\begin{tabular}{@{}lrrrr@{}}
\toprule
Metric & \shortstack{Direct\\SNN} & \shortstack{Haar\\events} & \shortstack{GAC\\SNN} & \shortstack{Continuous\\ANN} \\
\midrule
Accuracy (\%) & 92.731 & 91.141 & 92.277 & 90.290 \\
p50 (ms) & 52.624 & 63.501 & 54.615 & 31.892 \\
p95 (ms) & 53.339 & 68.632 & 55.248 & 32.667 \\
GPU/SoC (mJ) & 231.681 & 258.190 & 239.200 & 113.484 \\
CPU/CV (mJ) & 122.712 & 145.759 & 126.212 & 75.970 \\
SYS5V0 (mJ) & 255.925 & 304.103 & 262.897 & 141.228 \\
\bottomrule
\end{tabular}
\end{table}

\noindent\begin{minipage}{\columnwidth}
\subsection{Slip errors and checkpoint selection}
\label{app:tactile-object-protocols}

Screwdriver uses three training seeds, with two improving accuracy at epoch 40. Brush and SpoolSolder each use one paired seed and separate models trained for 40 epochs. Each protocol excludes its evaluation object from optimization and checkpoint selection; normalization and level-crossing thresholds use optimization data only. Metrics use overlapping windows. Table~\ref{tab:tactile-new-objects} compares the fixed and selected checkpoints.

\begin{table}[H]
\caption{Tactile evaluations, each excluding one object from training and selection. Rows compare one paired training seed on 30 held-out trials. Columns compare Direct with level crossings; accuracy, slip recall, and false-positive rate (FPR) are percentages over windows. Brush has 4,106 windows and SpoolSolder 9,890. Epoch 40 is fixed; the selected checkpoint minimizes cross-entropy, with the earliest epoch used in a tie. For brush and SpoolSolder, these minima occur at epochs 1--3. Models are trained separately from those for screwdriver.}
\label{tab:tactile-new-objects}
\centering\small
\setlength{\tabcolsep}{3pt}
\begin{tabular}{@{}lrrrr@{}}
\toprule
 & \multicolumn{2}{c}{Epoch 40} & \multicolumn{2}{c}{\shortstack{Selected\\checkpoint}} \\
\cmidrule(lr){2-3}\cmidrule(lr){4-5}
Metric (\%) & Direct & \shortstack{Level\\crossings} & Direct & \shortstack{Level\\crossings} \\
\midrule
\multicolumn{5}{l}{\textit{Brush}} \\
Accuracy & 59.21 & 70.85 & 75.40 & 70.58 \\
Slip recall & 64.96 & 62.51 & 43.92 & 50.15 \\
FPR & 42.60 & 26.54 & 14.74 & 23.03 \\
\addlinespace
\multicolumn{5}{l}{\textit{SpoolSolder}} \\
Accuracy & 50.71 & 49.54 & 58.38 & 56.38 \\
Slip recall & 86.56 & 63.56 & 54.14 & 78.77 \\
FPR & 59.60 & 54.49 & 40.40 & 50.06 \\
\bottomrule
\end{tabular}
\end{table}
\end{minipage}

\paragraph{Errors by slip state}
At epoch 40, screwdriver recall for stationary slip falls from 73.31\% with Direct to 49.37\% with level crossings, while recall for moving slip rises from 64.06\% to 67.18\%. On brush, the corresponding Direct and level-crossing recalls are 62.43\% and 54.14\% for stationary slip, and 66.45\% and 67.42\% for moving slip. On SpoolSolder they are 88.65\% and 60.33\% for stationary slip, and 84.53\% and 66.67\% for moving slip.

At checkpoints selected by minimum cross-entropy, recall for stationary slip is 46.96\% and 40.06\% for Direct and level crossings on brush, and 52.21\% and 85.89\% on SpoolSolder. Overall recall and recall for stationary slip can therefore favor different inputs.

\subsection{Matched device-cost comparisons}
\label{app:matched-cost-supplement}

Table~\ref{tab:additional-cost} summarizes the additional frozen-model replays. Each configuration uses eight 30-second blocks, 20 warm-up windows per block, batch size one, and 32-bit floating-point execution with TF32 disabled. Timing spans a standardized host window to returned host scores; energy is reported on the GPU/SoC rail using the scope in Section~\ref{sec:cost}.

\begin{table}[H]
\caption{Additional matched device-cost comparisons. Differences are encoder minus Direct: cached weighted Haar for MHEALTH, the continuous-coarse/event-detail hybrid for Widar, and level crossings otherwise. Entries are paired means $\pm$ sample standard deviations. Superscript s denotes variation across three seed-level paired means; b denotes variation across eight paired replay rounds of one seed. AC changes describe effective binary accumulations, not an energy estimate.}
\label{tab:additional-cost}
\centering\small
\setlength{\tabcolsep}{3pt}
\begin{tabular}{@{}lrr@{}}
\toprule
AC change & $\Delta$ latency (ms) & $\Delta$ rail (mJ) \\
\midrule
\multicolumn{3}{@{}l}{\textit{MHEALTH: 3 seeds, epoch 20, selection 96}} \\
$Higher$ & $8.240\pm0.054^{s}$ & $34.177\pm0.294^{s}$ \\
\addlinespace
\multicolumn{3}{@{}l}{\textit{Widar BVP: 1 seed, epoch 100, dev. 88}} \\
$+9.21\%$ & $12.476\pm0.072^{b}$ & $40.031\pm0.379^{b}$ \\
\addlinespace
\multicolumn{3}{@{}l}{\textit{Ninapro: 1 seed, epoch 40, day-3 dev. 80}} \\
$Higher$ & $0.141\pm0.052^{b}$ & $0.457\pm0.320^{b}$ \\
\addlinespace
\multicolumn{3}{@{}l}{\textit{uSkin screwdriver: 3 seeds, epoch 40, selection 64}} \\
$-40\%$ to $-37\%$ & $0.235\pm0.034^{s}$ & $0.813\pm0.084^{s}$ \\
\addlinespace
\multicolumn{3}{@{}l}{\textit{uSkin brush: 1 seed, epoch 40, selection 64}} \\
$-38.75\%$ & $0.203\pm0.138^{b}$ & $0.798\pm0.443^{b}$ \\
\addlinespace
\multicolumn{3}{@{}l}{\textit{uSkin SpoolSolder: 1 seed, epoch 40, selection 64}} \\
$-40.66\%$ & $0.241\pm0.163^{b}$ & $0.745\pm0.502^{b}$ \\
\addlinespace
\bottomrule
\end{tabular}
\end{table}

For MHEALTH, Figure~\ref{fig:inertial}c averages each model's block median over seeds; these displayed levels and the paired differences in Table~\ref{tab:additional-cost} are different summaries. Predictions retain their original classes. Replay operator counts show more effective accumulations for cached weighted Haar than for Direct.

The Widar device replay uses the seed-27102 epoch-100 weights and the same development input normalization as the recognition comparison; its device accuracies are 76.58\% and 77.41\%. For Ninapro, Figure~\ref{fig:semg} and its bootstrap interval use the recognition backend described in Section~\ref{sec:semg}. Full day-3 replay on the cost-measurement runtime yields user-averaged accuracies of 46.99\% and 48.40\%, with eight users improving, while some per-window predictions change. Its cost uses the same weights on 80 prespecified windows rather than all development windows.

For uSkin, recognition uses held-out objects, while these cost replays use prespecified selection windows from the same epoch-40 models. They therefore describe two evaluations of the same weights. Models selected by minimum selection loss have no corresponding cost measurement in this batch. Dense layer geometry and neuron-state sizes remain the same within each tactile pair; fewer effective accumulations do not imply that this dense backend skips the corresponding operations. The small paired differences are reported with their replay variability.

\end{document}